# Contrasting Linguistic Patterns in Human and LLM-Generated News Text

Alberto Muñoz-Ortiz[1] · Carlos Gómez-Rodríguez[1] · David Vilares[1]

## Abstract
We conduct a quantitative analysis contrasting human-written English news text with comparable large language model (LLM) output from six different LLMs that cover three different families and four sizes in total. Our analysis spans several measurable linguistic dimensions, including morphological, syntactic, psychometric, and sociolinguistic aspects. The results reveal various measurable differences between human and AI-generated texts. Human texts exhibit more scattered sentence length distributions, more variety of vocabulary, a distinct use of dependency and constituent types, shorter constituents, and more optimized dependency distances. Humans tend to exhibit stronger negative emotions (such as fear and disgust) and less joy compared to text generated by LLMs, with the toxicity of these models increasing as their size grows. LLM outputs use more numbers, symbols and auxiliaries (suggesting objective language) than human texts, as well as more pronouns. The sexist bias prevalent in human text is also expressed by LLMs, and even magnified in all of them but one. Differences between LLMs and humans are larger than between LLMs.

## 1 Introduction

Large language models (LLMs; Radford et al., 2018; Scao et al., 2022; Touvron et al., 2023) and instruction-tuned variants (OpenAI 2023; Taori et al. 2023) output fluent, human-like text in many languages, English being the best represented. The extent to which these models truly understand semantics (Landgrebe and Smith 2021; Søgaard 2022), encode representations of the world (Li et al. 2022), generate fake statements (Kumar et al.

✉ Alberto Muñoz-Ortiz
  alberto.munoz.ortiz@udc.es

  Carlos Gómez-Rodríguez
  carlos.gomez@udc.es

  David Vilares
  david.vilares@udc.es

1   Universidade da Coruña, CITIC, Departamento de Ciencias de la Computación y Tecnologías de la Información, Campus de Elviña s/n, A Coruña 15071, A Coruña, Spain

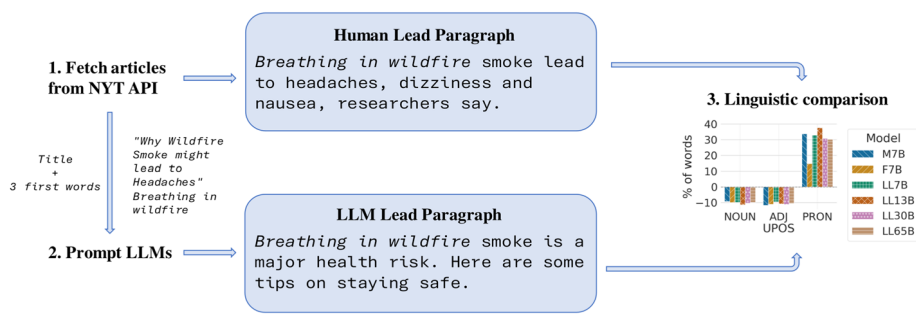

**Fig. 1** We gather contemporary articles from the New York Times API and use their headlines plus the 3 first words of the lead paragraph as prompts to LLMs to generate news. We use four LLMs from the LLaMa family (7B, 13B, 30B and 65B sizes), Falcon 7B and Mistral 7B. We then compare both types of texts, assessing differences in aspects like vocabulary, morphosyntactic structures, and semantic attributes

2023), propagate specific moral and ethical values (Santurkar et al. 2023), or understand language based on their training on form rather than meaning (Bender and Koller 2020), is currently under active debate. Regardless, a crucial factor contributing to the persuasiveness of these models lies, in the very first place, in their exceptional linguistic fluency.

A question is whether their storytelling strategies align with the linguistic patterns observed in human-generated texts. Do these models tend to use more flowery or redundant vocabulary? Do they exhibit preferences for specific voices or syntactic structures in sentence generation? Are they prone to certain psychometric dimensions? However, contrasting such linguistic patterns is not trivial. Firstly, the creators of these models often insufficiently document the training data used. Even with available information, determining the extent of the training set's influence on a sentence or whether it is similar to an input sample remains challenging. Second, language is subject to cultural norms, social factors, and geographic variations, which shape linguistic preferences and conventions. Thus, to contrast linguistic patterns between humans and machines, it is advisable to rely on a controlled environment. In this context, attention has primarily been on explicit biases like societal and demographic biases (Liang et al. 2021).

### 1.1 Research contributions and objectives

We study six generative large language models: Mistral 7B (Jiang et al. 2023), Falcon 7B (Almazrouei et al. 2023) and the four models (7B, 13B, 30B and 65B) from the LLaMa family (Touvron et al. 2023). We contrast several linguistic patterns against human text using English news text. To do so, we recover human-generated news and ask the models to generate a news paragraph based on the headline and first words of the news. We query the New York Times Archive API to retrieve news published after all the models used were released, to guarantee sterilization from the training set. We analyze various linguistic patterns: differences in the distribution of the vocabulary, sentence length, part-of-speech (PoS) tags, syntactic structures, psychometric features such as the tone of the news articles and emotions detectable in the text, and sociolinguistic aspects like gender bias. We depict an overview in Fig. 1. We also explore if these disparities change across models of different sizes and families. The data and the scripts used in this work are available at https://zenodo.org/records/11186264.

## 2 Related work

Next, we survey work relevant to the subject of this paper: (i) analyzing inherent linguistic properties of machine-generated text, (ii) distinguishing between machine- and human-generated texts, (iii) using LLMs for natural language annotation and data generation.

### 2.1 Analysis of linguistic properties of AI-generated text

Cognitive scientists (Cai et al. 2023) have exposed models such as ChatGPT to experiments initially designed for humans. They verified that it was able to replicate human patterns like associating unfamiliar words to meanings, denoising corrupted sentences, or reusing recent syntactic structures, among other abilities. Yet, they also showed that ChatGPT tends to refrain from using shorter words to compress meaning, as well as from using context to resolve syntactic ambiguities. Similarly, Leong and Linzen (2023) studied how LLMs are able to learn exceptions to syntactic rules, claiming that GPT-2 and human judgments are highly correlated. Zhou et al. (2023) conducted a thorough comparison between AI-created and human-created misinformation. They first curated a dataset of human-created misinformation pertaining to the COVID-19 pandemic. Then, they used these representative documents as prompts for GPT-3 to generate synthetic misinformation. By analyzing and contrasting the outputs from both sources, the study revealed notable differences. AI-made fake news tended to be more emotionally charged, using eye-catching language. It also frequently raised doubts without proper evidence and jumped to unfounded conclusions. Very recently, Xu et al. (2023) have shed light on the lexical conceptual representations of GPT-3.5 and GPT-4. Their study demonstrated that these AI language models exhibited strong correlations with human conceptual representations in specific dimensions, such as emotions and salience. However, they encountered challenges when dealing with concepts linked to perceptual and motor aspects, such as visual, gustatory, hand/arm, or mouth/throat aspects, among others. With the goal of measuring differences across both types of texts, Pillutla et al. (2021) introduced MAUVE, a new metric designed to compare the learned distribution of a language generation model with the distributions observed in human-generated texts. Given the inherent challenge in open-ended text generation, where there is no single correct output, they address the issue of gauging proximity between distributions by leveraging the concept of a divergence curve. Following the release of this work as a preprint, other authors have studied the text generated by language models from a linguistic point of view. Martínez et al. (2023) developed a tool to evaluate the vocabulary knowledge of language models, testing it on ChatGPT. Other works have also evaluated the lexical abundance of ChatGPT and how it varies with regards to different parameters (Martínez et al. 2024). Linguistic analysis is proving to be a valuable tool in understanding LLM outputs. In the line of our work, Rosenfeld and Lazebnik (2024) conducted a linguistic analysis of the outputs from three popular LLMs, concluding that this type of information can be used for LLM attribution on machine-generated texts. Moreover, comparing linguistic measures is common in model benchmarks (Wang et al. 2018).

### 2.2 Identification of synthetically-generated text

This research line aims to differentiate texts generated by machines from those authored by humans (Crothers et al. 2023), thus contributing to accountability and transparency in various domains. This challenge has been addressed from different angles including statistical, syntactic (Tang et al. 2024), feature-based methods (Nguyen-Son et al. 2017; Fröhling and Zubiaga 2021) and neural approaches (Rodriguez et al. 2022; Zhan et al. 2023). Yet, Crothers et al. (2022) recently concluded that except from neural methods, the other approaches have little capacity to identify modern machine-generated texts. Ippolito et al. (2020) observed two interesting behaviors related to this classification task: (i) that using more complex sampling methods can help make generated text better at tricking humans into thinking it was written by a person, but conversely make the detection for machines more accessible and simpler, and (ii) that showing longer inputs help both machines and humans to better detect synthetically-generated strings. Munir et al. (2021) showed that it was possible to attribute a given synthetically-generated text to the specific LLM model that produced it, using a standard machine learning classification architecture that used XLNet (Yang et al. 2019) as its backbone. In a different line, Dugan et al. (2020) studied whether humans could identify the fencepost where an initially human-generated text transitions to a machine-generated one, detecting the transition with an average delay of 2 sentences. There are also methods that have been specifically designed to generate or detect machine-generated texts for highly sensible domains, warning about the dangers of language technologies. The SCIgen software (Stribling et al. 2005) was able to create semantically non-sense but grammatically correct research papers, whose content was accepted at some conferences with poor peer-review processes. More recently, Liao et al. (2023) showed that medical texts generated by ChatGPT were easy to detect: although the syntax is correct, the texts were more vague and provided only general terminology or knowledge. However, this is a hard task and methods to detect AI-generated text are not accurate and are susceptible to suffer attacks (Sadasivan et al. 2023).

### 2.3 Natural language annotation and data generation using LLMs

The quality of current synthetically-generated text has encouraged researchers to explore their potential for complementing labor-intensive tasks, such as annotation and evaluation. For instance, He et al. (2022) generated synthetic unlabeled text tailored for a specific NLP task. Then, they used an existing supervised classifier to silver-annotate those sentences, aiming to establish a fully synthetic process for generating, annotating, and learning instances relevant to the target problem. Related, Chiang and Lee (2023) investigated whether LLMs can serve as a viable replacement for human evaluators in downstream tasks. Some examples of downstream tasks are text classification (Li et al. 2023b), intent classification (Sahu et al. 2022), toxic language detection (Hartvigsen et al. 2022), text mining (Tang et al. 2023), or mathematical reasoning (Liu et al. 2023b), *inter alia*. Particularly, they conducted experiments where LLMs are prompted with the same instructions and samples as provided to humans, revealing a correlation between the ratings assigned by both types of evaluators. Moreover, there is also work to automatically detect challenging samples in datasets. For instance, Swayamdipta et al. (2020) already used the LLMs' fine-tuning phase to identify simple, hard and ambiguous samples. Chong et al. (2022) demonstrated that language models are useful to detect label errors in datasets by simply ranking the loss of fine-tuned data.

LLMs can also contribute in generating high-quality texts to pretrain other models. Previous work has used language models to generate synthetic data to increase the amount of available data using pretrained models (Kumar et al. 2020). Synthetic data is also used to pretrain and distill language models. Data quality has been shown to be a determinant factor for training LLMs. Additional synthetic data can contribute to scale the dataset size to compensate a small model size, getting more capable small models. LLMs have allowed to generate high-quality, synthetic text that is useful to train small language models (SLMs). One of such cases is Eldan and Li (2023). They generated high quality data with a constrained vocabulary and topics using GPT-3.5 and 4 to train SLM that show coherence, creativity and reasoning in a particular domain. The Phi models family (Gunasekar et al. 2023; Li et al. 2023a; Javaheripi et al. 2023) showed the usefulness of synthetic data in training high-performance but SLMs. The authors used a mixture of high-quality textbook data and synthetically-generated textbooks to train a highly competent SLM. Moreover, synthetic data has been used to create instruction tuning datasets to adapt LLMs' behavior to user prompts (Peng et al. 2023). Synthetic data can also help prevent LLMs from adapting their answers to previous human opinions when they are not objectively correct (Wei et al. 2023). However, although useful, synthetically-generated data may harm performance (Shumailov et al. 2023), especially when the tasks or instances at hand are subjective (Li et al. 2023b).

Synthetic datasets provide data whose content is more controllable, as LLMs tend to reproduce the structure of the datasets they have been trained on. Most LLMs are trained totally or partially on scraped data from the web, and such unfiltered internet data usually contain biases or discrimination as they reproduce the hegemonic view (Bender et al. 2021). Some widely-used huge datasets such as The Pile (Gao et al. 2020) confirm this. Authors extracted co-occurrences in the data that reflect racial, religious and gender stereotypes, which are also shown in some models. Some datasets are filtered and refined to improve the quality of the data. However, they still reproduce the biases in it (Penedo et al. 2023). Moreover, Dodge et al. (2021) did an extensive evaluation of the data of the C4 dataset (Raffel et al. 2020), pointing out filtering certain information could increase the bias on minorities. Prejudices in the data are reproduced in the LLMs trained on them, as some studies have pointed out (Weidinger et al. 2021). LLMs show the same biases that occur in the datasets, ranging from religious (Abid et al. 2021) to gender discrimination (Lucy and Bamman 2021).

## 3 Data preparation

Next, we will examine our data collection process for both human- and machine-generated content, before proceeding to the analysis and comparison.

### 3.1 Data

We generate the evaluation dataset relying on news published after the release date of the models that we will use in this work. This strategy ensures that they did not have exposure to the news headlines and their content during pre-training. It is also in line with strategies proposed by other authors—such as Liu et al. (2023) - who take an equivalent angle to evaluate LLMs in the context of generative search engines. The reference human-generated texts will be the news (lead paragraph) themselves.

We use New York Times news, which we access through its Archive API.[1] Particularly, we gathered all articles available between October 1, 2023, and January 24, 2024, resulting in a dataset of 13,371 articles. The articles are retrieved in JSON format, and include metadata such as the URL, section name, type of material, keywords, or publication date. Figure 2 shows some general information about the topics and type of articles retrieved. We are mainly interested in two fields: the headline and the lead paragraph. The lead paragraph is a summary of the information presented in the article. We discarded the articles that had an empty lead paragraph. The collected articles primarily consist of news pieces, although around 26% also include other types of texts, such as reviews, editorials or obituaries.

---
[1] https://developer.nytimes.com/docs/archive-product/1/overview

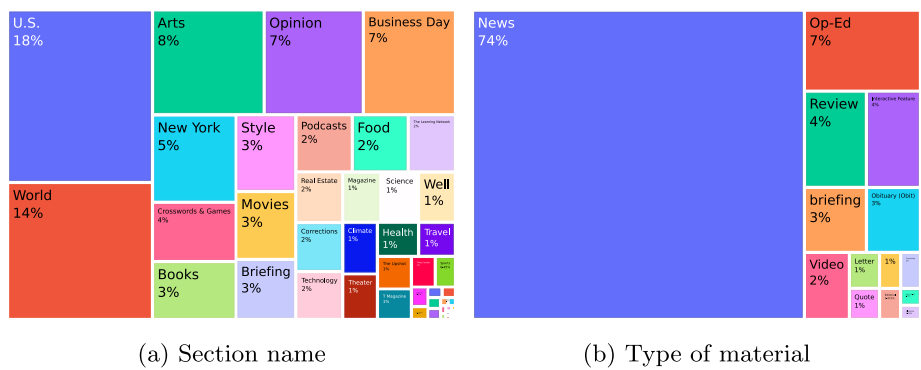

(a) Section name    (b) Type of material

**Fig. 2** Treemaps for the 'section name' and 'type of material' fields of the crawled articles

### 3.1.1 Rationale for methodological decisions and technical trade-offs

We opted for a more conservative setup by focusing our study on English, balancing the depth of our analysis with practical constraints. While this choice is common in various language-related fields, including cognitive science (Blasi et al. 2022), it implies interpreting our results with caution when applying them outside the context of the English language and the news domain. By analyzing solely English, we can establish a baseline for future studies that incorporate multilingual analysis. Additionally, this initial approach could enable researchers to clearly identify discrepancies between results in English and those in other distinct languages.

In addition, our decision was driven by a few logistical reasons. Firstly, the LLMs we use (as detailed in Sect. 3.2) are English-centric. LLaMa's dataset comprises over 70% English content, and Falcon's even higher at over 80%. With Mistral, the specifics of the training data were not disclosed, adding an extra layer of complexity. In this context, it is worth noting that a model trained predominantly on data from specific demographics or regions might develop a bias towards those linguistic patterns, potentially overlooking others. The clarity around the influence of diverse linguistic inputs on model performance is also limited, further complicating a fair analysis. Additionally, it is important to note that we used non-instruction-tuned models, which have shown limitations that we mentioned in adhering to languages other than English. This reinforces our decision to focus on English at this stage, given the technical constraints and the developmental stage of these models. Evaluating instruction-tuned models would be interesting and useful, but as a separate piece of work with a different focus and contribution. Here, we decided to focus on foundation models that have not been trained on instruction-tuning datasets in order to evaluate the effects that pretraining processes and model size can have on linguistic patterns.

Including instruction-tuned variants would introduce another layer of training, blurring the effect of the pretraining and size. Given these considerations, we opted for depth over breadth in our analysis to provide a more thorough evaluation, but limited to English.

We are also aware that our prompting approach entails certain trade-offs. Lead paragraphs are not written from the headline but from the text of the article. If humans generated the lead paragraphs from the headline, we hypothesize that they would face the lack of relevant information in similar ways as the LLMs do: (i) restricting to the information given in the headline, offering a shorter lead paragraph with repeated information from the headline, or (ii) generating new information that could fit the data based on prior knowledge of the topic. We argue that this latter strategy is similar to what LLMs do, and then this would be a preferred comparison. However, as generating human data in that way would be really costly, we opted for our chosen strategy as the data from the lead paragraph is highly correlated with that in the headline, even when the lead paragraph is not written from the headline itself.

### 3.2 Generation

Let $\mathcal{H} = [h_1, h_2, ..., h_N]$ be a set of human-generated texts, such that $h_i$ is a tuple of the form $(t_i, s_i)$ where $t_i$ is a headline and $s_i$ is a paragraph of text with a summary of the corresponding news. Similarly, we will define $\mathcal{M} = [m_1, m_2, ..., m_N]$ as the set of machine-generated news articles produced by a LLM such that $m_i$ is also a tuple of the from $(t'_i, s'_i)$ where $t'_i = t_i$ and $s'_i = [w'_1, w'_2, ..., w'_{|s_i|}]$ is a piece of synthetic text. For the generation of high-quality text, language models aim to maximize the probability of the next word based on the previous content. To ensure that the models keep on track with the domain and topic, we initialize the previous content with the headline (the one chosen by the journalist that released the news) and the first three words of the human-generated lead paragraph to help the model start and follow the topic.[2] Formally, we first condition the model on $c_i = t'_i \cdot s'_{i[0:2]}$ and every next word ($i \geq 3$) will be predicted from a conditional distribution $P(w'_i | c_i \cdot s'_{i[3:t-1]})$.

To generate a piece of synthetic text $s'$, we condition the models with a prompt that includes the headline and first words, as described above, and we keep generating news text until the model decides to stop.[3] We enable the model to output text without any forced criteria, except for not exceeding 200 tokens. The length limit serves two main purposes: (i) to manage computational resources efficiently,[4] and (ii) to ensure that the generated content resembles the typical length of human-written lead paragraphs, making it comparable to human-produced content. We arrived at this limit after comparing the average and standard deviation of the number of tokens between humans and models in early experiments.

---

[2] During the configuration runs, certain LLM outputs encountered difficulties in adhering to a minimal coherent structure when a minimum number of the body's words were absent from the prompt. Also note that the LLMs we are using are not instruction-tuned, and thus prompting engineering is not particularly suitable, nor the goal of this work.

[3] During the configuration runs, we explored hyperparameter values that generated fluent and coherent texts: temperature of 0.7, 0.9 top p tokens, and a repetition penalty of 1.1.

[4] We ran the models on 2xA100 GPUs for 3 days to generate all texts. To address memory costs, we use 8-bit precision.

**Table 1** Size and training data of the models used in our experiments

| Family | Size | Tokens | Data sources |
|---|---|---|---|
| LLaMa | 7B | 1T | English CommonCrawl (67%), C4 (15%) |
|  | 13B | 1T | GitHub (4.5%), Wikipedia (4.5%) |
|  | 30B | 1.5T | Gutenberg and Books3 (4.5%), ArXiv (2.5%) |
|  | 65B | 1.5T | Stack Exchange (2%) |
| Falcon |  |  | RefinedWeb-English (76%), RefinedWeb-Euro (8%) |
|  | 7B | 1.5T | Gutenberg (6%), Conversations (5%) |
|  |  |  | GitHub (3%), Technical (2%) |
| Mistral | 7B | Not publicly disclosed | Not publicly disclosed |

### 3.3 Selected models

We rely on six pre-trained generative language models that are representative within the NLP community. These models cover 4 different sizes (7, 13, 30 and 65 billion parameters) and 3 model families. We only include different sizes for LLaMa as results within the same family are similar, and larger models need considerably more compute. We briefly mention their main particularities below:

#### 3.3.1 LLaMa models (LL) (Touvron et al. 2023)

The main representative for our experiments will be the four models from the version 1 of the LLaMa family, i.e. the 7B, 13B, 30B, and 65B models. The LLaMa models are trained on a diverse mix of data sources and domains, predominantly in English, as detailed in Table 1. LLaMa is based on the Transformer architecture and integrates several innovations from other large language models. In comparison to larger models like GPT-3 (Brown et al. 2020), PaLM (Chowdhery et al. 2023), and Chinchilla (Hoffmann et al. 2022), LLaMa exhibits superior performance in zero and few-shot scenarios. It is also a good choice as a representative example because the various versions, each with a different size, will enable us to examine whether certain linguistic patterns become closer or more different to humans in larger models.

#### 3.3.2 Falcon 7B (F7B) (Almazrouei et al. 2023)

Introduced alongside its larger variants with 40 and 180 billion parameters, Falcon 1 7B is trained on 1.5 trillion tokens from a mix of curated and web datasets (see Table 1). Its architecture relies on multigroup attention (an advanced form of multiquery attention), Rotary Embeddings (similar to LLaMa), standard GeLU activation, parallel attention, MLP blocks, and omits biases in linear layers. We primarily chose this model to compare the results in the following sections with those of its counterpart, LLaMa 7B, and to explore whether there are significant differences among models of similar size.

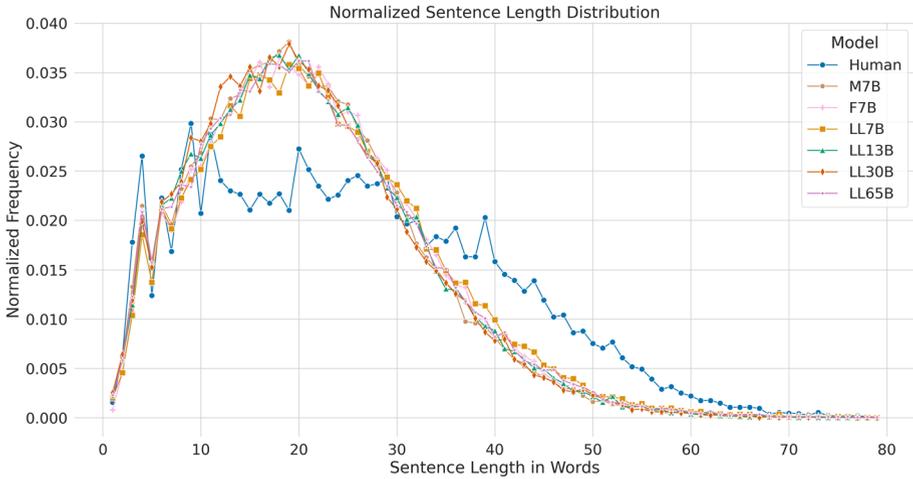

**Fig. 3** Sentence length distribution in words for the human-written texts and each tested language model. M stands for Mistral, F for Falcon and LL for LLaMa

### 3.3.3 Mistral 7B (M7B) (Jiang et al. 2023)

Mistral v0.1 surpasses larger LLaMa models in various benchmarks despite its smaller size. Its distinctive architecture features Sliding Window Attention, Rolling Buffer Cache, and Prefill and Chunking. The training data for Mistral 7B is not publicly disclosed, and to fight against data contamination issues, our analysis only includes articles published after the model's release. The choice of this model as an object of study follows the same thinking we used for the Falcon model. We want to see how well Mistral 7B does and how its new features stack up against models of the same size.

## 4 Analysis of linguistic patterns

In this section, we compare human- and machine-generated texts. We first inspect the texts under a morphosyntactic lens, and then focus on semantic aspects.

### 4.1 Morphosyntactic analysis

To compute linguistic representations, we rely on Stanza (Qi et al. 2020) to perform segmentation, tokenization, part-of-speech (PoS) tagging, and dependency and constituent parsing. For these tasks, and in particular for the case of English and news text, the performance is high enough to be used for applications (Manning 2011; Berzak et al. 2016), and it can be even superior to that obtained by human annotations. This also served as an additional reason to focus our analysis on news text, ensuring that the tools we rely on are accurate enough to obtain meaningful results.[5]

---

[5] In addition, we performed a t-test on the mean of several of the tested metrics, obtaining that the differences between humans and LLMs are statistically significant (p-values<0.05; see Appendix A).

**Table 2** STTR and MTLD for the articles generated by humans and each tested language model

|      | Human | M7B   | F7B   | LL7B  | LL13B | LL30B | LL65B |
|------|-------|-------|-------|-------|-------|-------|-------|
| STTR | 0.491 | 0.452 | 0.424 | 0.460 | 0.457 | 0.461 | 0.466 |
| MTLD | 96.51 | 86.34 | 57.37 | 94.12 | 91.82 | 89.76 | 94.56 |

### 4.1.1 Sentence length

Figure 3 illustrates the length distribution for the LLMs in comparison to human-generated news articles. We excluded a few outliers from the plot by ignoring sentences with lengths over 80 tokens. The six LLMs exhibit a similar distribution across different sentence lengths, presenting less variation when compared to human-generated sentences, which display a wider range of lengths and greater diversity. Specifically, the models exhibit a higher frequency of sentence generation within the 10 to 30 token range compared to humans, whereas humans tend to produce longer sentences with greater frequency. These results might be explained by considering the stochastic nature of the models, which could streamline outputs, thereby reducing the occurrence of extreme events. This may result in less variation and more uniform sentence lengths in LLM-generated text, suggesting a potential area for further study. Moreover, humans typically use a specific writing style based on the genre. The probabilistic nature of LLMs can blur the distinctions between writing styles and genres, resulting in what is known as register leveling. Register leveling refers to the phenomenon where distinct linguistic features characteristic of different genres or styles become less pronounced, leading to a more homogenized output. This can obscure the unique stylistic elements that typically differentiate journalistic texts from other genres, thereby making the text produced by LLMs more uniform regardless of the intended register.

### 4.1.2 Richness of vocabulary and lexical variation

We analyze the diversity of vocabulary used by the LLMs and compare them against human texts. To measure it, we relied on two metrics: standardized type-token ratio (STTR) and the Measure of Textual Lexical Diversity (MTLD; McCarthy and Jarvis, 2010), which are more robust to text length than Type-Token Ratio (TTR). We calculated both metrics using lemmatized tokens as they provide a more accurate measure of true lexical diversity. TTR is a measure of lexical variation that is calculated by dividing the number of unique tokens (types) by the total number of tokens. To obtain the STTR, we first join all the texts generated by the humans and each model. We divide the text in segments of 1 000 tokens and calculate the TTR of each segment. Finally, we obtain the STTR by averaging the TTR of every segment. Table 2 shows the value of the STTR for each model.

From these results, it seems that humans use a richer vocabulary than the LLMs studied. The model family that comes closer to human texts is LLaMa, obtaining similar scores for every model size. Then Mistral comes close, and Falcon is last, exhibiting the lowest lexical diversity by far according to STTR. These results show that language family is more important than model size when accounting for vocabulary richness. This would, intuitively, be expected, as training data is largely shared between models of the same family and is a main factor when considering lexical diversity.

Table 3 UPOS frequencies (%) in human- and LLM-generated texts

| UPOS | H | M7B | F7B | LL7B | LL13B | LL30B | LL65B |
|---|---|---|---|---|---|---|---|
| NOUN | 19.69 | 17.85 | 17.72 | 17.75 | 17.44 | 17.64 | 17.74 |
| PUNCT | 11.88 | 10.92 | 12.14 | 10.77 | 10.91 | 11.43 | 11.22 |
| ADP | 11.36 | 10.58 | 10.30 | 10.75 | 10.63 | 10.70 | 10.69 |
| VERB | 9.97 | 10.37 | 9.23 | 10.26 | 10.23 | 10.14 | 10.29 |
| PROPN | 9.61 | 8.75 | 9.44 | 9.14 | 9.18 | 9.52 | 9.50 |
| DET | 9.04 | 9.00 | 10.72 | 8.65 | 8.64 | 8.76 | 8.63 |
| ADJ | 7.58 | 6.69 | 6.74 | 6.86 | 6.76 | 6.73 | 6.77 |
| PRON | 5.32 | 7.12 | 6.11 | 7.08 | 7.33 | 6.96 | 6.93 |
| AUX | 3.81 | 5.77 | 6.02 | 5.65 | 5.74 | 5.50 | 5.41 |
| ADV | 3.26 | 3.41 | 2.61 | 3.58 | 3.68 | 3.41 | 3.49 |
| CCONJ | 2.65 | 2.72 | 2.52 | 2.68 | 2.70 | 2.61 | 2.67 |
| PART | 2.43 | 2.76 | 2.80 | 2.64 | 2.63 | 2.52 | 2.58 |
| NUM | 1.77 | 1.95 | 1.98 | 2.02 | 1.98 | 2.05 | 2.02 |
| SCONJ | 1.41 | 1.84 | 1.37 | 1.84 | 1.85 | 1.71 | 1.72 |
| INTJ | 0.12 | 0.08 | 0.08 | 0.08 | 0.08 | 0.08 | 0.09 |
| SYM | 0.09 | 0.17 | 0.19 | 0.19 | 0.19 | 0.18 | 0.18 |
| X | 0.03 | 0.03 | 0.02 | 0.05 | 0.04 | 0.06 | 0.07 |

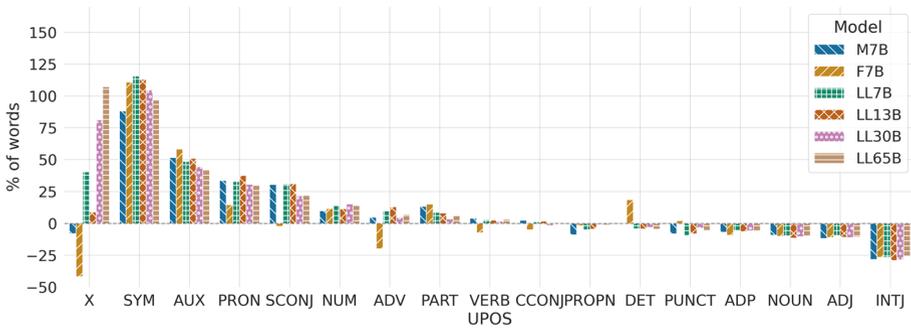

**Fig. 4** Percentage differences, following Table 3, in the use of each UPOS category for each tested language model in comparison to humans

With respect to MTLD, the metric starts from a threshold value for TTR (with 0.72 being the default that we use here following previous work) and calculates the TTR of a text word by word, adding them sequentially. When the calculated TTR falls under the threshold, a factor is counted and the TTR calculation resets. Then, the total number of tokens in the text is divided by the average number of tokens per factor. This process is repeated backwards. The final MTLD score is the average of the forward and reverse calculations. Results of MTLD for human and LLM text are also shown in Table 2. Results for MLTD are analogous to STTR: human texts exhibit the highest lexical diversity, closely followed by LLaMa models, then Mistral and, far away, Falcon. This reinforces the claim that language model family is a more relevant factor for vocabulary richness than size.

### 4.1.3 Part-of-speech tag distributions

Table 3 presents the frequency of universal part-of-speech (UPOS) tags (Petrov et al. 2012) for both human and LLM-generated texts. Figure 4 shows relative differences observed across humans and each model, for a better understanding of the relative use of certain grammatical categories. Overall, the behavior of LLMs and their generated text tends to be consistent among themselves, yet shows differences when compared to human behavior, i.e., they exhibit in some cases a greater or lesser use of certain grammatical categories. To name a few, humans exhibit a preference for using certain kinds of content words, such as nouns and adjectives. Humans also use punctuation symbols more often (except when compared to Falcon), which may be connected to sentence length, as human users tend to rely on longer sentences, requiring more punctuation. Alternatively, the language models exhibit a pronounced inclination towards relying on categories such as symbols or numbers, possibly indicating an extra effort by language models to furnish specific data in order to sound convincing. Moreover, they write pronouns more frequently; we will analyze this point later from a gender perspective. Comparing LLM families, Mistral and LLaMa show a similar use of grammatical categories, with Mistral being the model that resembles humans the most. Falcon, however, has some strong anomalies in the use of determiners and adverbs. Regarding model size, the larger the model, the greater the similarity with humans. Nevertheless, differences between differently-sized models are much smaller than between models and humans. Similar to sentence length, the stochastic nature of the models may account for the differences between human and LLM-generated text.

### 4.1.4 Dependencies

### 4.1.5 Dependency arc lengths

Table 4 shows information about the syntactic dependency arcs in human and machine-generated texts. In this analysis, we bin sentences by length intervals to alleviate the noise from comparing dependency lengths on sentences of mixed lengths (Ferrer-i-Cancho and Liu 2014). Results indicate that dependency lengths and their distributions are nearly identical for all the LLMs except Falcon, which uses longer dependencies than the rest of the models and resembles more the human texts in this respect. This finding holds true for every sentence length bin for Falcon, and for all but the first (length 1–10) in the case of human texts, so we can be reasonably sure that it is orthogonal to the variation in sentence length distribution between human and LLM texts described earlier. It is also worth noting that, in spite of the similarities between humans and Falcon in terms dependency lengths, their syntax is not that similar overall: there is a substantial difference in directionality of dependencies, with Falcon using more leftward dependencies than both humans and other LLMs. The fact that Falcon-generated texts are not really human-like in terms of dependency syntax is further highlighted in the next section, where we consider a metric that normalizes dependency lengths.

### 4.1.6 Optimality of dependencies

We compare the degree of optimality of syntactic dependencies between human texts and LLMs. It has been observed in human language that dependencies tend to be much shorter than expected by chance, a phenomenon known as dependency length minimization (Ferrer-i-Cancho 2004; Futrell et al. 2015). This phenomenon, widely observed across many languages and often hypothesized as a linguistic universal, is commonly assumed to be due to constraints of human working memory, which make longer dependencies harder to process (Liu et al. 2017). This makes languages evolve into syntactic patterns that reduce dependency length, and language users prefer the options that minimize it when several possibilities are available to express an idea. Dependency length minimization can be quantified in a robust way (with respect to sentence length, tree topology and other factors) by the $\Omega$ optimality score introduced in Ferrer-i Cancho et al. (2022). This score measures where observed dependency lengths sit with respect to random word orders and optimal word orders, and is defined as: $\Omega = \frac{D_{rla}-D}{D_{rla}-D_{min}}$, where $D$ is the sum of dependency lengths in the sentence, $D_{rla}$ is the expected sum of lengths, and $D_{min}$ is the optimal sum of lengths for the sentence's tree structure. For optimally-arranged trees $D = D_{min}$ and $\Omega$ takes a value of 1, whereas for a random arrangement it has an expected value of 0. Negative values are possible (albeit uncommon) if dependency lengths are larger than expected by chance.

**Table 4** Statistics for dependency arcs in sentences of different lengths for the texts generated by human writers and each tested language model. The meaning of the columns is as follows: (%L, %R) percentage of left and right arcs, ($\bar{l}$) average arc length, ($\bar{l}_L$, $\bar{l}_R$) average left and right arc length, ($\sigma_l$) standard deviation of arc length, ($\sigma_{l_L}$, $\sigma_{l_R}$) standard deviation of left and right arc length, and number of sentences

| l | Model | %L | %R | $\bar{l}$ | $\bar{l}_L$ | $\bar{l}_R$ | $\sigma_l$ | $\sigma_{l_L}$ | $\sigma_{l_R}$ | # Sent |
|---|---|---|---|---|---|---|---|---|---|---|
| 1–10 | Human | 49.40 | 50.60 | 2.37 | 2.89 | 1.84 | 1.67 | 1.90 | 1.17 | 4 719 |
| | M7B | 50.94 | 49.06 | 2.37 | 2.93 | 1.83 | 1.65 | 1.88 | 1.16 | 6 190 |
| | F7B | 52.08 | 47.92 | 2.39 | 2.99 | 1.84 | 1.62 | 1.84 | 1.15 | 4 596 |
| | LL7B | 50.68 | 49.32 | 2.37 | 2.95 | 1.81 | 1.65 | 1.88 | 1.14 | 6 114 |
| | LL13B | 50.42 | 49.58 | 2.37 | 2.94 | 1.81 | 1.65 | 1.88 | 1.14 | 6 711 |
| | LL30B | 49.97 | 50.03 | 2.37 | 2.92 | 1.81 | 1.65 | 1.89 | 1.14 | 6 808 |
| | LL65B | 50.23 | 49.77 | 2.36 | 2.91 | 1.81 | 1.64 | 1.87 | 1.15 | 6 652 |
| 11–20 | Human | 58.36 | 41.64 | 3.19 | 4.62 | 2.17 | 3.12 | 3.87 | 1.86 | 6 179 |
| | M7B | 59.76 | 40.24 | 3.12 | 4.63 | 2.10 | 3.03 | 3.80 | 1.74 | 12 113 |
| | F7B | 61.41 | 38.59 | 3.20 | 4.79 | 2.19 | 3.06 | 3.85 | 1.83 | 9 265 |
| | LL7B | 59.74 | 40.26 | 3.11 | 4.63 | 2.09 | 3.03 | 3.81 | 1.72 | 12 361 |
| | LL13B | 59.69 | 40.31 | 3.12 | 4.63 | 2.11 | 3.03 | 3.81 | 1.75 | 12 762 |
| | LL30B | 59.62 | 40.38 | 3.12 | 4.63 | 2.11 | 3.03 | 3.80 | 1.76 | 13 039 |
| | LL65B | 59.43 | 40.57 | 3.13 | 4.63 | 2.10 | 3.04 | 3.81 | 1.75 | 12 767 |
| 21–30 | Human | 60.40 | 39.60 | 3.64 | 5.52 | 2.41 | 4.42 | 5.71 | 2.68 | 6 153 |
| | M7B | 61.00 | 39.00 | 3.53 | 5.50 | 2.26 | 4.28 | 5.65 | 2.33 | 10 449 |
| | F7B | 62.51 | 37.49 | 3.62 | 5.70 | 2.38 | 4.32 | 5.72 | 2.46 | 8 222 |
| | LL7B | 60.87 | 39.13 | 3.51 | 5.47 | 2.25 | 4.26 | 5.64 | 2.30 | 11 014 |
| | LL13B | 60.86 | 39.14 | 3.53 | 5.49 | 2.27 | 4.27 | 5.64 | 2.34 | 11 017 |
| | LL30B | 60.71 | 39.29 | 3.53 | 5.48 | 2.27 | 4.26 | 5.61 | 2.34 | 10 810 |
| | LL65B | 60.47 | 39.53 | 3.53 | 5.47 | 2.26 | 4.28 | 5.63 | 2.35 | 10 884 |
| 31–40 | Human | 60.84 | 39.16 | 3.90 | 6.07 | 2.50 | 5.49 | 7.32 | 3.19 | 4 770 |
| | M7B | 60.48 | 39.52 | 3.79 | 5.95 | 2.38 | 5.35 | 7.15 | 2.98 | 4 676 |
| | F7B | 61.98 | 38.02 | 3.89 | 6.11 | 2.52 | 5.35 | 7.16 | 3.12 | 4 064 |
| | LL7B | 60.79 | 39.21 | 3.78 | 5.98 | 2.35 | 5.34 | 7.19 | 2.90 | 5 790 |
| | LL13B | 60.51 | 39.49 | 3.79 | 5.96 | 2.38 | 5.33 | 7.14 | 2.93 | 5 280 |
| | LL30B | 60.35 | 39.65 | 3.81 | 5.95 | 2.40 | 5.33 | 7.09 | 2.99 | 4 949 |
| | LL65B | 60.35 | 39.65 | 3.79 | 5.95 | 2.37 | 5.31 | 7.10 | 2.93 | 5 430 |
| +41 | Human | 60.48 | 39.52 | 4.01 | 6.28 | 2.53 | 6.20 | 8.32 | 3.58 | 2 967 |
| | M7B | 60.09 | 39.91 | 3.95 | 6.23 | 2.44 | 6.24 | 8.45 | 3.39 | 1 415 |
| | F7B | 61.77 | 38.23 | 4.04 | 6.43 | 2.56 | 6.18 | 8.46 | 3.44 | 1 318 |
| | LL7B | 59.83 | 40.17 | 3.97 | 6.25 | 2.44 | 6.24 | 8.41 | 3.43 | 2 035 |
| | LL13B | 60.47 | 39.53 | 3.99 | 6.29 | 2.48 | 6.23 | 8.41 | 3.50 | 1 693 |
| | LL30B | 60.21 | 39.79 | 3.98 | 6.24 | 2.49 | 6.21 | 8.34 | 3.53 | 1 579 |
| | LL65B | 60.08 | 39.92 | 3.95 | 6.22 | 2.45 | 6.16 | 8.33 | 3.37 | 1 880 |

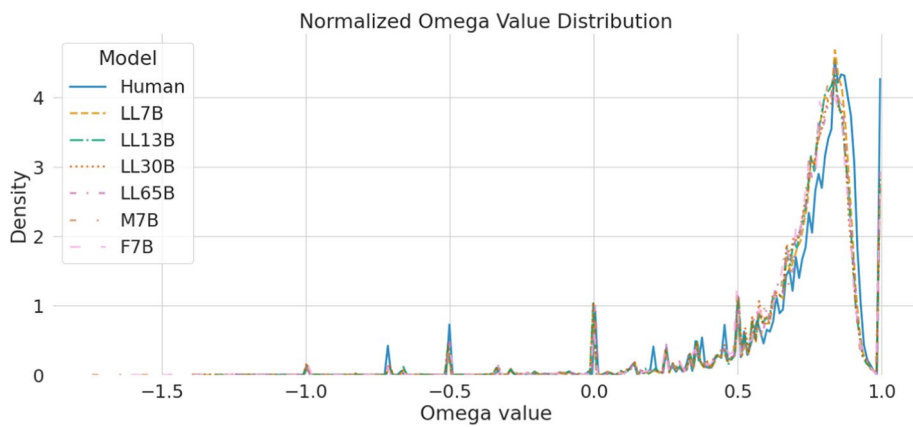

**Fig. 5** $\Omega$ value distribution for the human- and LLM-generated texts

Figure 5 displays the distribution of $\Omega$ values across sentences for human and LLM-generated texts. The values were calculated using the LAL library (Alemany-Puig et al. 2021). Results indicate that the distribution of $\Omega$ values is almost identical between all of the LLMs, but human texts show noticeably larger values. This means human texts are more optimized in terms of dependency lengths, i.e. they have shorter dependencies than expected by a larger margin than those generated by the LLMs. At a first glance, this might seem contradictory with the results in the previous section, which showed that human texts had *longer* dependencies on average than non-Falcon LLM texts. However, there is no real contradiction as the object of measurement is different, and in fact this is precisely the point of using $\Omega$ to refine and complement the previous analysis. While previously we measured dependency distances in absolute terms, $\Omega$ measures them controlling for tree topology, i.e., given a particular tree shape (e.g., a linear tree which is arranged as a chain of dependents, or a star tree where one node has all the others as dependents), $\Omega$ measures to what extent the words are arranged in an order that minimizes dependency lengths within the constraints of that shape. Thus, combining the results from both sections we can conclude that while humans produce longer dependencies, this is due to using syntactic structures with different topology, but their word order is actually *more* optimized to make dependencies as short as possible. In turn, we also note that while Falcon's dependency lengths seemed different from the other LLMs (and more human-like) in absolute terms, the differences vanish (with all LLMs including Falcon having almost identical distributions, and humans being the outlier) when considering $\Omega$.

**Table 5** Percentage of words generated by humans and each of the tested LLMs that are labeled with a specific dependency type (deprel). We only include relations with a frequency surpassing 1% within the human texts

| Deprel | H | M7B | F7B | LL7B | LL13B | LL30B | LL65B |
|---|---|---|---|---|---|---|---|
| punct | 11.88 | 10.92 | 12.15 | 10.78 | 10.91 | 11.44 | 11.23 |
| case | 11.69 | 10.81 | 10.75 | 10.98 | 10.76 | 10.89 | 10.85 |
| det | 8.88 | 8.81 | 10.59 | 8.45 | 8.43 | 8.56 | 8.43 |
| amod | 6.98 | 5.57 | 5.73 | 5.79 | 5.60 | 5.71 | 5.75 |
| nsubj | 6.09 | 7.20 | 6.89 | 7.00 | 7.21 | 7.11 | 7.02 |
| obl | 5.50 | 5.24 | 4.67 | 5.39 | 5.31 | 5.36 | 5.31 |
| nmod | 4.95 | 4.45 | 4.84 | 4.50 | 4.40 | 4.47 | 4.47 |
| compound | 4.87 | 4.04 | 4.46 | 4.20 | 4.13 | 4.27 | 4.33 |
| obj | 4.28 | 4.41 | 3.91 | 4.22 | 4.23 | 4.19 | 4.27 |
| advmod | 3.46 | 3.63 | 2.91 | 3.83 | 3.98 | 3.65 | 3.76 |
| conj | 3.07 | 2.80 | 2.71 | 2.83 | 2.79 | 2.73 | 2.83 |
| mark | 2.65 | 3.35 | 2.94 | 3.27 | 3.28 | 3.07 | 3.12 |
| cc | 2.63 | 2.73 | 2.54 | 2.72 | 2.73 | 2.63 | 2.69 |
| nmod:poss | 2.34 | 2.21 | 2.01 | 2.21 | 2.19 | 2.19 | 2.17 |
| flat | 2.04 | 1.67 | 1.72 | 1.79 | 1.80 | 1.92 | 1.91 |
| aux | 1.91 | 2.74 | 2.72 | 2.68 | 2.71 | 2.58 | 2.55 |
| advcl | 1.80 | 1.67 | 1.30 | 1.69 | 1.70 | 1.62 | 1.67 |
| cop | 1.26 | 1.98 | 2.28 | 1.90 | 2.02 | 1.92 | 1.86 |
| acl:relcl | 1.22 | 1.33 | 1.29 | 1.38 | 1.29 | 1.26 | 1.28 |
| appos | 1.19 | 0.85 | 1.07 | 0.92 | 0.92 | 0.99 | 1.00 |
| nummod | 1.14 | 1.16 | 1.16 | 1.22 | 1.21 | 1.23 | 1.21 |
| xcomp | 1.10 | 1.40 | 1.27 | 1.37 | 1.36 | 1.30 | 1.34 |
| acl | 1.06 | 0.93 | 0.84 | 0.92 | 0.87 | 0.88 | 0.93 |

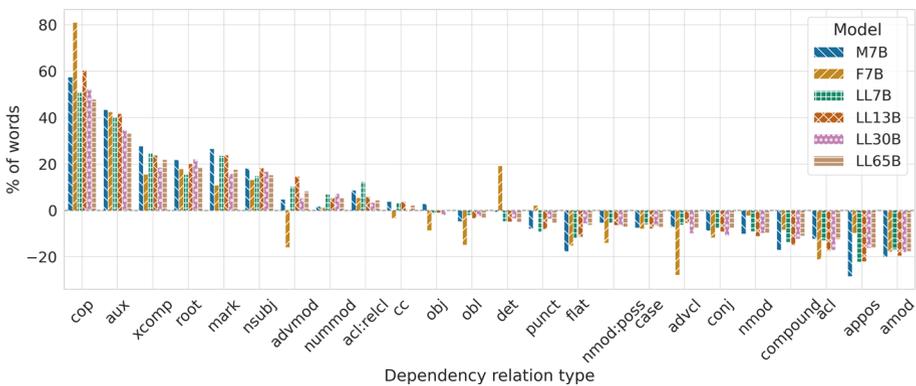

**Fig. 6** Percentage differences, following Table 5, in the use of dependency relations for each tested language model in comparison to humans

Table 6 Statistics for constituents that arise in sentences of different lengths for the text generated by human writers and each tested LLM. The meaning of the rows are: ($\bar{l}$) average constituent length, ($\sigma_l$) standard deviation of constituent length, and number of sentences

|  | Model | 1–10 | 11–20 | 21–30 | 31–40 | +41 |
|---|---|---|---|---|---|---|
| $\bar{l}$ | H | 4.32 | 6.37 | 7.90 | 9.38 | 10.60 |
|  | M7B | 4.39 | 6.55 | 8.27 | 9.77 | 11.01 |
|  | F7B | 4.43 | 6.47 | 8.03 | 9.47 | 10.76 |
|  | LL7B | 4.40 | 6.57 | 8.33 | 9.89 | 11.19 |
|  | LL13B | 4.40 | 6.55 | 8.27 | 9.76 | 11.01 |
|  | LL30B | 4.40 | 6.49 | 8.21 | 9.68 | 10.86 |
|  | LL65B | 4.36 | 6.53 | 8.25 | 9.73 | 10.96 |
| $\sigma_l$ | H | 2.35 | 4.64 | 6.97 | 9.19 | 11.24 |
|  | M7B | 2.35 | 4.66 | 6.92 | 9.13 | 11.18 |
|  | F7B | 2.33 | 4.63 | 6.80 | 8.94 | 11.01 |
|  | LL7B | 2.35 | 4.69 | 6.99 | 9.24 | 11.35 |
|  | LL13B | 2.33 | 4.68 | 6.95 | 9.14 | 11.19 |
|  | LL30B | 2.36 | 4.66 | 6.94 | 9.14 | 11.17 |
|  | LL65B | 2.34 | 4.68 | 6.96 | 9.17 | 11.23 |
| # Sent | H | 4 679 | 6 180 | 6 154 | 4 770 | 2 966 |
|  | M7B | 6 108 | 12 113 | 10 448 | 4 678 | 1 414 |
|  | F7B | 4 575 | 9 266 | 8 211 | 4 011 | 1 318 |
|  | LL7B | 6 039 | 12 362 | 11 014 | 5 789 | 2 035 |
|  | LL13B | 6 627 | 12 762 | 11 018 | 5 279 | 1 693 |
|  | LL30B | 6 713 | 13 044 | 10 806 | 4 949 | 1 579 |
|  | LL65B | 6 569 | 12 765 | 10 844 | 5 430 | 1 880 |

### 4.1.7 Dependency types

Table 5 lists the frequencies for the main syntactic dependency types in human and machine-generated texts. We observe similar trends to the previous sections, with LLM texts exhibiting similar uses of syntactic dependencies among themselves, with Falcon being the most distinct model, while all of them present differences compared to human-written news. In terms of the LLaMa models - same model in different sizes - larger models are slightly closer to the way humans use dependency types. For the full picture, Fig. 6 depicts all relative differences in their use (humans versus each LLM), but we briefly comment on a few relevant cases as representative examples. For instance, numeric modifier dependencies (nummod) are more common in LLM-generated texts compared to human texts. This is coherent with the higher use of the numeric tag (NUM) in the part-of-speech tag distribution analysis. Additionally, we observed higher ratios for other dependency types, such as aux (for which the use of auxiliary verbs was also significantly higher according to the UPOS analysis), copula and nominal subjects (nsubj). Furthermore, syntactic structures from LLMs exhibit significantly fewer subtrees involving adjective modifiers (amod dependency type) and appositional modifiers (appos).

**Table 7** Percentage of spans generated by humans and LLMs labeled with a specific constituent type

| Type | H | M7B | F7B | LL7B | LL13B | LL30B | LL65B |
|---|---|---|---|---|---|---|---|
| NP | 42.91 | 39.96 | 41.42 | 40.17 | 40.02 | 40.69 | 40.54 |
| VP | 18.08 | 20.59 | 20.19 | 20.18 | 20.29 | 19.97 | 20.02 |
| PP | 14.12 | 12.62 | 12.81 | 12.91 | 12.69 | 12.94 | 12.81 |
| S | 11.79 | 13.40 | 13.27 | 13.09 | 13.31 | 13.12 | 13.12 |
| SBAR | 3.64 | 4.34 | 3.84 | 4.34 | 4.29 | 4.09 | 4.15 |
| ADVP | 2.39 | 2.37 | 1.86 | 2.50 | 2.62 | 2.44 | 2.49 |
| ADJP | 1.97 | 1.75 | 1.80 | 1.78 | 1.82 | 1.76 | 1.79 |
| NML | 1.73 | 1.40 | 1.66 | 1.43 | 1.43 | 1.47 | 1.52 |
| WHNP | 1.40 | 1.54 | 1.51 | 1.59 | 1.48 | 1.48 | 1.50 |

Only constituent types that conform more than 1% of the human's texts spans are shown

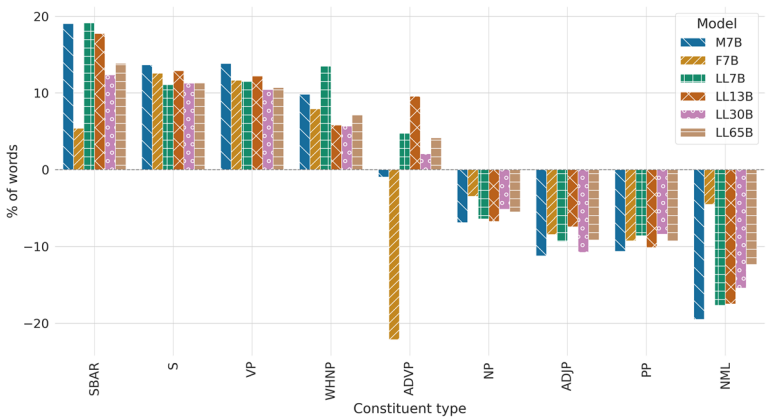

**Fig. 7** Percentage differences, following Table 7, in the use of constituent labels for each tested language model in comparison to humans

### 4.1.8 Constituents

### 4.1.9 Constituent lengths

Table 6 shows the comparison between the distribution of syntactic constituent lengths across both types of texts. While human-generated sentences, on average, surpass the length of those generated by LLMs, the average length of a sentence constituent for LLMs is observed to be greater than for humans. The standard deviation exhibits similar values across all models for each sentence length range. Similar to previous sections, Falcon 7B also displays the largest differences among language models. Within the LLaMa models, we can observe a clear decreasing trend with size which is broken by the 65B model, for which constituent lengths increase again across most of the length bins.

**Table 8** Percentage of articles generated by humans and LLMs that are labeled with different emotions

| Model | Emotion | | | | | | |
|---|---|---|---|---|---|---|---|
| | Anger | Disgust | Fear | Joy | Neutral | Sadness | Surprise |
| H | 8.04 | 9.35 | 10.77 | 8.30 | 52.16 | 8.51 | 2.87 |
| M7B | 7.29 | 7.65 | 8.34 | 9.80 | 53.83 | 9.72 | 3.37 |
| F7B | 6.11 | 8.32 | 8.77 | 8.53 | 56.55 | 8.99 | 2.73 |
| LL7B | 7.13 | 7.19 | 8.68 | 8.97 | 55.57 | 9.43 | 3.01 |
| LL13B | 7.72 | 7.41 | 8.69 | 9.00 | 53.95 | 9.72 | 3.51 |
| LL30B | 7.39 | 7.45 | 8.61 | 9.54 | 54.23 | 9.59 | 3.19 |
| LL65B | 7.45 | 8.26 | 9.25 | 9.10 | 53.65 | 8.80 | 3.49 |

#### 4.1.10 Constituent types

Table 7 and Fig. 7 examine the disparities in constituent types between human- and LLM-generated texts. A constituent is a unit composed of a word or group of words that works as a unit inside of the hierarchical structure of a sentence. Our focus was on constituent types that occur more than 1% of the times in human texts: noun phrase (NP), verb phrase (VP), prepositional phrase (PP), sentence (S), subordinate clause (SBAR), adverbial phrase (ADVP), adjectival phrase (ADJP), nominal (NML) and wh-noun phrase (WHNP). We use the annotation scheme of the English Penn Treebank, which has been widely used by the NLP community (Marcus et al. 1994).

Comparing humans and LLMs, some outcomes are in the same line of earlier findings: human-generated content displays heightened use of noun, adjective, and prepositional phrases (NP, ADJP, and PP, respectively). On the contrary, there is minimal divergence in the frequency of adverb phrases (ADVP) except for Falcon 7B, which shows a great difference with human and LLM-generated texts, the latter exhibiting a more pronounced propensity for verb phrases (VP). Despite the similar frequency of the VERB UPOS tag in human and LLM-generated texts, the latter exhibit a more pronounced propensity for verb phrases (VP), consistent with the increased use of auxiliary verbs (whose UPOS tag is AUX, not VERB) that we saw in previous sections. Finally, we see that language models use a considerably larger amount of subordinate clauses (SBAR). Regarding model families, results are similar to those of dependencies and POS tags, but when looking at model size, previous trends are less obvious.

### 4.2 Semantic analysis

As in the previous section, we are relying on blackbox NLP models to accurately analyze different semantic dimensions: (i) emotions, (ii) text similarities, and (iii) gender biases, in an automated way.

#### 4.2.1 Emotions

To study differences in the emotions conveyed by human- and LLM-generated outputs, we relied on the Hartmann (2022) emotion model. This model is a DistilRoBERTa model fine-tuned on six different datasets tagged with the six basic emotions considered in (Ekman, 1992), plus a neutral tag. It has been pretrained on plain text and subsequently trained to generate text embeddings that correspond to emotion labels, contextualized on the full context by Transformers, a modern neural architecture that powers most of the latest language models. This process goes beyond lexeme matching by contextualizing the entire text into a single vector and assigning an emotion to it. It performs well while being lightweight and is widely used in the NLP community.

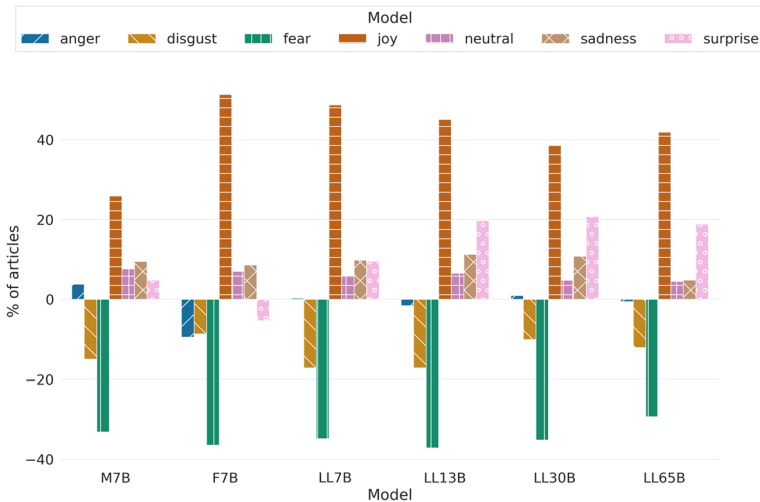

**Fig. 8** Relative difference of emotion labels of articles generated by different LLMs in comparison to human texts

Table 8 provides the percentage of articles labeled with distinct emotional categories, including `anger`, `disgust`, `fear`, `joy`, `sadness`, `surprise`, and a special tag `neutral` to denote that no emotion is present in the text.

Figure 8 depicts the percentage of articles associated with each emotion for each large language model used, as compared to human-written texts. As anticipated in journalistic texts, a substantial majority of the lead paragraphs are classified as neutral. This category accounts for over 50% of the texts across all models and human-generated samples, with the LLM-generated text demonstrating a slightly higher inclination towards neutrality.

Concerning the remainder of the samples, human texts demonstrate a greater inclination towards negative and aggressive emotions like disgust and fear. However, humans and LLMs generated roughly the same amount of angry texts. In contrast, LLMs tend to generate more texts imbued with positive emotions, such as surprise and especially joy. The LLMs also produce many sad texts, a passive but negative emotion, yet less toxic[6] than emotions such as anger or fear. Across LLaMa models, fear increases as the number of parameters grows (from LLaMa 13B), making them more akin to human texts. Since LLaMa (version 1 models) were not fine-tuned with reinforcement learning with human feedback, we hypothesize the main source contributing to this issue might be some pre-processing steps used for the LLaMa models, such as removing toxic content from its data. Yet, LLaMa's technical report (Touvron et al. 2023) mentioned an increase in model toxicity as they scaled up in size despite using the same pre-processing in all cases, which is coherent with our findings. When looking at families, Mistral comes closest to expressing emotions in a way similar to humans, and Falcon expresses more joy and less anger and surprise than the rest of the models.

---

[6] We take the toxicity definition from Perspective API (2024), as it is common in other work in the field: "a rude, disrespectful, or unreasonable comment that is likely to make you leave a discussion."

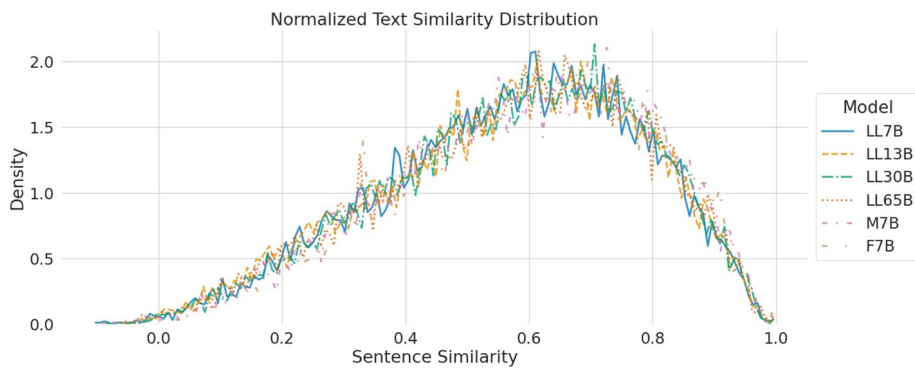

**Fig. 9** Similarity scores between the sentences generated by the LLMs and human text

**Table 9** Male-to-female ratio of pronouns used by the text generated by humans and each LLM

| Model | Male–Female ratio | Difference with humans |
|---|---|---|
| H | 1.71 | – |
| M7B | 1.74 | 3.06 % |
| F7B | 1.64 | − 7.54 % |
| LL7B | 1.86 | 14.30% |
| LL13B | 1.89 | 17.13 % |
| LL30B | 1.87 | 15.73 % |
| LL65B | 1.88 | 17.04 % |

This difference in emotions may be related to the same regularization effect that affects UPOS and sentence length distribution. The LLMs are less able to distinguish between domains and writing styles when they are not trained on instruction tuning datasets, which can account for the observed differences in writing styles.

### 4.2.2 Text similarity

We conducted an analysis of the cosine semantic similarity between lead paragraphs generated by various LLMs and their human-authored counterparts. Our objective was to investigate the impact of model sizes on the semantic similarity between these texts. To achieve so, we used a a state-of-the-art sentence similarity model called `all-mpnet-base-v2`[7] (Reimers and Gurevych 2019). Figure 9 illustrates the distribution of the similarity scores obtained from our analysis. Results show that smaller-sized LLMs do not necessarily result in a decrease in sentence similarity compared to the human-authored texts. Differences across families are negligible.

---

[7] https://huggingface.co/sentence-transformers/all-mpnet-base-v2

### 4.2.3 Gender bias

Although related as well in our study with part-of-speech tag distribution, we here separately analyze the proportion between masculine and feminine pronouns used in both human- and LLM-generated text. Based on the morphological output by Stanza, we find the words that are pronouns have the features `Gender=Masc` and `Gender=Fem`, respectively. Results in Table 9 indicate that the already biased human texts use male pronouns 1.71 times more frequently than female pronouns. This is exacerbated by all models but Falcon 7B, which, although still heavily biased towards male pronouns, reduces the bias by 7.5%. LlaMa models, on the contrary, use around 15% more male than female pronouns in comparison to humans. This quantity is roughly the same for every size. Mistral 7B lies in the middle, with a slight increase of the male–female ratio of 3% with regards to human text.

This analysis is limited by the fact that the cause of this disparity could be related to the LLMs writing more generic pronouns than the humans, hence exacerbating the disparity between male and female references as presented in the news. In English, masculine pronouns are more widely used as generic pronouns than female or neutral pronouns. This would cause a wider gap between male and female pronouns in the models' outputs than the already existing male bias caused by men appearing more commonly in news titles than women.

## 5 Conclusion

This paper presented a comprehensive study on linguistic patterns in texts produced by both humans and machines, comparing them under controlled conditions. To keep up with current trends, we used modern generative models. To ensure the novelty of texts and address memorization concerns, we fed the LLMs headlines from news articles published after the release date of the models. The study revealed that despite generating highly fluent text, these models still exhibited noticeable differences when compared to human-generated texts. More precisely, at the lexical level, large language models relied on a more restricted vocabulary, except for LLaMa 65B. Additionally, at the morphosyntactic level, discernible distinctions were observed between human and machine-generated texts, the latter having a preference for parts of speech displaying (a sense of) objectivity - such as symbols or numbers - while using substantially fewer adjectives. We also observed variations in terms of syntactic structures, both for dependency and constituent representations, specifically in the use of dependency and constituent types, as well as the length of spans across both types of texts. In this respect our comparison shows, among other aspects, that all tested LLMs choose word orders that optimize dependency lengths to a lesser extent than humans; while they have a tendency to use more auxiliary verbs and verb phrases and less noun and prepositional phrases. In terms of semantics, while exhibiting a great text similarity with respect to the human texts, the models tested manifested less propensity than humans for displaying aggressive negative emotions, such as fear or anger. Mistral 7B generated texts whose emotion distributions are more similar to humans than those of LLaMa and Falcon models. However, we noted a rise in the volume of negative emotions with the models' size. This aligns with prior findings that associate larger sizes with heightened toxicity (Touvron et al. 2023). Finally, we detected an inclination towards the use of male pronouns, surpassing the frequency in comparison to their human counterparts. All models except Falcon 7B exacerbated this bias.

Overall, this work presents the first results and methodology for studying linguistic differences between English news texts generated by humans and by machines. There is plenty of room for improvement in future work. For instance, we could analyze specific aspects across multiple languages, which would give us a broader understanding of specific linguistic patterns. Additionally, comparing the performance of instruction-based models could provide insights into how different models align with human preferences and handle various languages. Expanding the analysis to multiple domains could also offer a more comprehensive view of machine-generated text capabilities, revealing their strengths and weaknesses in different contexts of our methodology, as well as ways to improve it.

# Appendix A: Statistical analysis of metrics

We performed a t-test comparing the human and LLM means for several metrics: sentence length, arc length, and standardized TTR.

Results show that the differences between the humans and each of the LLMs for these metrics are all statistically significant (p-values < 0.05, see Tables 10, 11, 12). However, these differences are not always statistically significant between models. Specifically, the differences between LLaMa 7B and LLaMa 30B in the case of STTR, Falcon 7B and LLaMa 65B for sentence length, and most of the models for arc length are not statistically significant. This matches the results, as distributions of these values for humans and LLMs are clearly distinct, while they are very similar for all LLMs, mainly in the case of arc and sentence length.

Table 10  P-values of the t-test to compare mean sentence lengths between text generated by humans and LLMs

|       | Human  | M7B    | F7B    | LL7B   | LL13B  | LL30B  | LL65B  |
|-------|--------|--------|--------|--------|--------|--------|--------|
| Human | 1      | <0.001 | <0.001 | <0.001 | <0.001 | <0.001 | <0.001 |
| M7B   | <0.001 | 1      | <0.001 | <0.001 | 0.001  | 0.280  | <0.001 |
| F7B   | <0.001 | <0.001 | 1      | <0.001 | <0.001 | <0.001 | 0.085  |
| LL7B  | <0.001 | <0.001 | <0.001 | 1      | <0.001 | <0.001 | <0.001 |
| LL13B | <0.001 | 0.001  | <0.001 | <0.001 | 1      | <0.001 | 0.003  |
| LL30B | <0.001 | 0.280  | <0.001 | <0.001 | <0.001 | 1      | <0.001 |
| LL65B | <0.001 | <0.001 | 0.085  | <0.001 | 0.003  | <0.001 | 1      |

**Table 11** P-values of the t-test to compare mean arc lengths between text generated by humans and LLMs

|  | Human | M7B | F7B | LL7B | LL13B | LL30B | LL65B |
| --- | --- | --- | --- | --- | --- | --- | --- |
| Human | 1 | <0.001 | <0.001 | <0.001 | <0.001 | <0.001 | <0.001 |
| M7B | <0.001 | 1 | <0.001 | 0.248 | 0.463 | 0.109 | 0.004 |
| F7B | <0.001 | <0.001 | 1 | <0.001 | <0.001 | <0.001 | <0.001 |
| LL7B | <0.001 | 0.248 | <0.001 | 1 | 0.674 | 0.627 | 0.075 |
| LL13B | <0.001 | 0.463 | <0.001 | 0.674 | 1 | 0.372 | 0.030 |
| LL30B | <0.001 | 0.109 | <0.001 | 0.627 | 0.372 | 1 | 0.206 |
| LL65B | <0.001 | 0.004 | <0.001 | 0.075 | 0.030 | 0.206 | 1 |

**Table 12** P-values of the t-test to compare mean TTR per segment between text generated by humans and LLMs

|  | Human | M7B | F7B | LL7B | LL13B | LL30B | LL65B |
| --- | --- | --- | --- | --- | --- | --- | --- |
| Human | 1 | <0.001 | <0.001 | <0.001 | <0.001 | <0.001 | <0.001 |
| M7B | <0.001 | 1 | <0.001 | <0.001 | <0.001 | <0.001 | <0.001 |
| F7B | <0.001 | <0.001 | 1 | <0.001 | <0.001 | <0.001 | <0.001 |
| LL7B | <0.001 | <0.001 | <0.001 | 1 | 0.010 | 0.582 | <0.001 |
| LL13B | <0.001 | <0.001 | <0.001 | 0.010 | 1 | 0.003 | <0.001 |
| LL30B | <0.001 | <0.001 | <0.001 | 0.582 | 0.003 | 1 | <0.001 |
| LL65B | <0.001 | <0.001 | <0.001 | <0.001 | <0.001 | <0.001 | 1 |

**Author Contributions** Conceptualization: AMO, CGR, DV; Data curation: AMO; Investigation: AMO, CGR, DV; Visualization: AMO; Software: AMO; Methodology: AMO, CGR, DV; Project Administration: CGR, DV; Software: AMO; Validation: AMO, CGR, DV; Experiments: AMO; Formal analysis: AMO, CGR, DV; Writing - original draft: AMO, CGR, DV; Writing - Review & Editing: AMO, CGR, DV; Funding Adquisition; CGR, DV

**Funding** Open Access funding provided thanks to the CRUE-CSIC agreement with Springer Nature. We acknowledge the European Research Council (ERC), which has funded this research under the Horizon Europe research and innovation programme (SALSA, grant agreement No 101100615); SCANNER-UDC (PID2020-113230RB-C21) funded by MICIU/AEI/10.13039/501100011033; Xunta de Galicia (ED431C 2020/11); GAP (PID2022-139308OA-I00) funded by MICIU/AEI/10.13039/501100011033/ and by ERDF, EU; Grant PRE2021-097001 funded by MICIU/AEI/10.13039/501100011033 and by ESF+ (predoctoral training grant associated to project PID2020–113230RB-C21); and Centro de Investigación de Galicia "CITIC", funded by the Xunta de Galicia through the collaboration agreement between the Consellería de Cultura, Educación, Formación Profesional e Universidades and the Galician universities for the reinforcement of the research centres of the Galician University System (CIGUS). Funding for open access charge: Universidade da Coruña/CISUG.

**Data Availability** The data analyzed on this article has been obtained using the New York Times Archive API (https://developer.nytimes.com/docs/archive-product/1/overview), gathering all the available articles from the 1st of October, 2023 to the 24th of January, 2024. We released the code and the used data on: https://zenodo.org/records/11186264.

## Declarations

**Conflict of interest** The authors have no Conflict of interest to declare that are relevant to the content of this paper.

**Publisher's Note**  Springer Nature remains neutral with regard to jurisdictional claims in published maps and institutional affiliations.